\documentclass{ewic-latex}
\usepackage{url}
\usepackage{hyperref}

\begin{document}

\runningheads{Axelsson}{Speculative Design of Equitable Robotics: Queer Fictions and Futures}

\conference{}

\title{Speculative Design of Equitable Robotics: Queer Fictions and Futures}

\authorone{Minja Axelsson\\
University of Cambridge\\
UK\\
\email{mwa29@cam.ac.uk}}

\begin{abstract}
This paper examines the speculative topic of equitable robots through an exploratory essay format. It focuses specifically on robots by and for LGBTQ+ populations. It aims to provoke thought and conversations in the field about what aspirational queer robotics futures may look like, both in the arts and sciences. First, it briefly reviews the state-of-the-art of queer robotics in fiction and science, drawing together threads from each. Then, it discusses queering robots through three speculative design proposals for queer robot roles: 1) reflecting the queerness of their “in-group” queer users, building and celebrating “in-group” identity, 2) a new kind of queer activism by implementing queer robot identity performance to interact with “out-group” users, with a goal of reducing bigotry through familiarisation, and 3) a network of queer-owned robots, through which the community could reach each other, and distribute and access important resources. The paper then questions whether robots should be queered, and what ethical implications this raises. Finally, the paper makes suggestions for what aspirational queer robotics futures may look like, and what would be required to get there.
\end{abstract}


\maketitle

\section{Introduction: Questions Explored and Opened in this Essay}
What are queer robots? How might we design queer robots, and should we? What would an aspirational queer robotics future look like? This explorative essay starts to feel out answers to these questions, presenting a starting point for discussion by the community. It introduces some of the examinations of queer robots in both science and the arts, and through this identifies three speculative design proposals for meaningful queer robotics. It then questions the desirability of queer robots and the related ethical issues, and discusses queer robots and robotics practice as potential tools for challenging heteronormativity. 

The work aims to explore what equitable robotic futures could and should look like from a queer lens. Researchers have called for more equitable practices in the human-robot interaction (HRI) field, in order to centre social justice by conducting research which articulates which communities it aims to serve and how (\cite{zhu2024robots}), and to identify who benefits from the designed robots (\cite{ostrowski2022ethics}). Recently, researchers have also focused on the ethical and socio-technical questions surrounding different applications of human-robot interaction (\cite{{axelsson2025four}}), as well as used design fiction to examine the values embedded in dystopian and utopian speculative robot designs (\cite{axelsson2025trash}). This work aims build on these works to identify what ethical robotics futures and speculative robot designs could look like for and with the LGBTQ+ population, who are currently under-represented in robotics.

\section{The State-of-the-Art of Queer Robotics in Fiction and Science}

It is worth examining both the arts and science when investigating the state-of-the-art of queer robots. As scientific scholarship on queer robots is nascent, exploring imaginaries documented in the arts can help us understand current visions in queer robotics. Fiction serves an important role in inspiring potential futures, and shapes research and innovation (\cite{bassett2013better}). This makes the discussion of both fiction and science related to queer robots relevant.

In fiction, robots are often ``queer'' by default---genderless and asexual. However, these forms of queerness are not often meaningfully engaged with or identified as queerness. On the other hand, some fictional depictions of robots have been definitively and explicitly queer. A relatively recent prominent example is the human-like woman-presenting robot Niska, from the British TV show Humans\footnote{\url{https://www.imdb.com/title/tt4122068/}, Accessed 06-06-2025}. Niska is unapologetic and headstrong, qualities that also characterise her queer romance, which centres woman-presenting characters in the historically male-centred genre of science fiction. As an android, Niska is on a journey of self-discovery---a fitting metaphor for the queer experience. Unfortunately, her human-likeness in behaviour and especially appearance serves to somewhat simplify what could otherwise be a rich exploration of robotic, woman-centred queerness. 

Another recent example is Ann Leckie’s Imperial Radch series\footnote{\url{https://annleckie.com/novel/ancillary-justice/}, Accessed 06-06-2025}, which queers gender in a novel way for the science fiction genre. While not strictly a robot, the main character Breq is an AI, embodied in a single human. Functionally, Breq is a flesh robot with an AI brain. In the story, Breq historically inhabited multiple bodies of multiple genders, experiencing a multiplicity of queerness in both gender and sexuality. As a result, Breq refers to all other characters as she/her, and does not easily perceive differences in gender. Instead, Breq guesses based on cues gleaned from the environment: clothing, how other people relate to a particular person, manner of speaking, etc. This shows interesting parallels to how people decipher gender as we go about the world, sometimes so automatically that it is not given a second thought, while Breq has to actively make these gender perception decisions. Leckie's story raises questions about whether robots should be designed to interpret similar social signals---and what would the implications of that be? A meaningful experience of recognition for under-served gender and sexuality groups, or sinister surveillance that can be used for nefarious purposes? 

Though fiction and the arts are particularly good vehicles for speculating on queer identities in robotics, these representations are few and far between. The same can be seen in scientific robotics research: works exploring robot identity are just beginning to emerge. Recently, researchers have started to explore how trans and non-binary people view social robots (\cite{stolp2024more}). Some (e.g., \cite{winkle2021flexibility}) have examined the concept of identity performance in social robots. Social robots are robots that interact socially with humans through e.g., speech, gestures and facial expressions. \cite{winkle2021flexibility} propose that identity performance in social robotics could be used to ``model alternative identity norms'' and ``reflect an actively feminist, anti-racist and pro social justice design and development stance''. Key ideas in the nascent field of queer robotics are the fluidity and customisability of robot identity (\cite{winkle2021flexibility}), as well as the multiplicity (\cite{seaborn2022identified}) of robot identities.

\section{Queering Robots through Speculative Design}

Building on social identity theory, Seaborn argues that robots may even “pass” as certain identity groups, giving the particular example of robots passing as human (\cite{seaborn2022identified}). The vocabulary of ``passing'' in the queer community is most typically used with reference to trans experiences. This starts to introduce challenging questions of whether robots can ``pass'' as belonging to a specific gender identity (be that cis, trans, woman, man, non-binary, or other gender identities). It also introduces questions of whether robots can pass as members of a sexual identity group. In its simplest binary formation, these groups would be queer and non-queer, or in other words heteronormative and non-heteronormative. However, such groups can also be defined in varied, intersectional ways.

If one were to build robots accepting that robots could legitimately express queer identities with regards to gender and/or sexuality, potential speculative design use cases for queer robots could be: 

\begin{enumerate}
    \item \textbf{Reflecting the queerness of their ``in-group'' queer users}: These robots could express a queer identity, in order to reflect, build and celebrate “in-group” identity with queer users. 
    
    \textbf{User base}: These robots could be especially helpful for younger queer users, who are just beginning to recognise and explore their queer identity. 
    
    \textbf{Aimed outcome of use}: The robot could be a tool for building self-esteem as a new member of the community, and introduce elements of queer culture.
    
    \item \textbf{A new kind of queer activism and advocacy}: These robots would use queer identity performance to interact with ``out-group'' users (i.e., straight and cisgender people). 
    
    \textbf{User base}: The robot could be especially useful for non-queer and/or heteronormative people with limited exposure to queer communities, who may hold prejudiced beliefs but are willing to engage with an open mind to challenge their assumptions.
    
    \textbf{Aimed outcome of use}: These robots would have the goal of reducing bigotry through familiarisation, without a queer person having to serve the role of a patient educator. The robot could present elements of queer history, and educate on the importance and benefits of inclusion.
    
    \item \textbf{A network of queer-owned robots}, through which the community could reach each other, and distribute and access resources of importance to the community. 

    \textbf{User base}: Queer people who want to build community (i.e., keep in touch with existing queer community members and support new community members) through technological tools, including robots.
    
    \textbf{Aimed outcome of use}: These robots would not need to express a queer identity, but would serve queer users by giving information about e.g., queer community meet-ups, gender transition, sexual health, etc. These robots would function more as communicative tools rather than as social others, in order to connect queer users and help them build community. 
\end{enumerate}

As an additional feature to these proposed speculative designs, the multiplicity (as discussed by \cite{seaborn2022identified}) of robot identities could allow robots to hold multiple and even conflicting queer identities at a time, fluidly moving from one identity to the other. Speculative design and fictional explorations of multiplicitous robot identities could include, e.g., non-binary robots, or pan- and bisexual robots. Could such multiplicity allow for the meaningful exploration of, e.g., non-binary robot identities in fiction? At the moment, non-binary identities are under-represented in fiction, and queer robots could provide an avenue for these explorations.

However, it should be noted that in the wrong hands, gender-queering fictional robots could allow a vehicle for less reflective writers to bring in queer characters without having to fully engage with their lived experience and material realities. As \cite{{stolp2024more}} find in their work exploring trans and non-binary perspectives on social robots, the dehumanisation of non-cis people is a risk when designing non-cis robot identities. At its worst, this could turn ``queer'' robots (both in terms of gender and sexuality) into a convenient box-ticking exercise. However, if done sensitively by and with queer people, such fictional and speculative design explorations of queer robots could be a deeply meaningful way of exploring gender and sexuality, as a tool for reflection of the human world. Such explorations could lead into research on meaningful queer robot identities in the sciences. These are only some of the important ethical considerations raised by the conceptualisation of queer robots, more of which are explored in the next section.

\section{Should we Queer Robots? Some Ethical Considerations}

Representations of queer robots in the arts and fiction could serve as alternative, justice-centred visions for the future. Current representations are often stereotyped and heteronormative. Prominent examples of this are the movies Ex Machina\footnote{\url{https://www.imdb.com/title/tt0470752/}, Accessed on 06-06-2025} and Companion\footnote{\url{https://www.imdb.com/title/tt26584495/}, Accessed  on 06-06-2025}, in which a sexualised and woman-presenting robot murders the somewhat hapless man-presenting main character. To challenge these normative visions, there is a need for stories with queerer examinations of these traditional narratives of power, love, and death. Such stories could expand imaginaries of robots: How would an identifiably (rather than as a by-default setting for a robot) asexual or agender robot move through the world? How would they be perceived by others? What could these queer robots tell us about the living of queer lives?

In science and research, introducing robots that express queer identities raises ethical considerations. One concern is that queering robot identities could serve as an avenue for rainbow capitalism (see e.g., \cite{falco2020rainbow})---i.e., using queer identity in a transactional manner merely to drive further profits, while not engaging in serious inclusive efforts. However, queer robot identity could introduce meaningful applications for robots. Could queer robot identities provide LGBTQ+ people with some useful reflection, a feeling of relating, or a support for exploring identity? Could it provide cis and straight people with ``access'' to non-heteronormative identities, to normalise these identities without a queer person having to sacrifice their time, energy and patience to serve as an educator? Or, would this just be wearing queerness as a mask and a costume, and function as a form of appropriation? Can robots ever have an ``authentic'' queer (or non-queer) identity? These questions I also pose to the community to discuss further.

One argument for queering robots could be that robots are already designed according to certain identity norms. In fact, human-shaped robots have been designed to be particularly sexist (see, e.g., \cite{robertson2010gendering}), enforcing rigid feminine and masculine gender norms in function and appearance. For instance, the voice of voice assistant Siri was originally designed to capitulate to sexual harassment (\cite{bergen2016blush}), reinforcing heteronormative systems of power and oppression. ``I’d blush if I could'' said Siri, when called one of the slurs uniquely reserved for women. From this point of view, choosing to design explicitly queer robots could be used to challenge existing and unconscious heteronormative assumptions of gender and sexuality, which are currently embedded into robot design. Researchers have advocated for this, and have pointed out the lack of robots designed for queer people, and the lack of research on the impacts of robotics on queer people (see \cite{korpan2024queer}). 

As such, if we do not queer robots, the robots may remain heteronormative, in potentially unethical and harmful ways that reinforce gendered stereotypes. This provides an argument that the ethical choice would be to queer robots. In fact, \cite{miranda2023identity} find that robotics has not yet meaningfully engaged with intersectionality in robot design. They argue that ``human-robot interaction has the power to influence human norms and culture'', and as such should meaningfully engage with under-represented identities. While queering robots introduces potential ethical issues of rainbow capitalism -driven transactionality, not queering robots is already resulting in ethical issues, in which robots continue replicating and perpetuating oppressive heteronormative roles and narrow identity narratives. In an aspirational future, robots that have been ascribed and successfully express (i.e., ``pass'' with) a queer identity could serve as tools for social and even political inclusion, liberation, empowerment and justice for queer people.

\section{An Aspirational Queer Robotics Future}

Having discussed all of this, what would an aspirational, queer robotics future look like?

Perhaps it will include the design and research of robots with multiplicitous queer identities, which would interact both with “in-group” and “out-group” people. This will require funding and resources to research how robots can be queered, and how they impact queer people. This in turn requires  the recognition of the value of exploring how robotics can serve under-served demographics. An aspirational future could also include fiction that features queer robots, challenging heteronormative assumptions present in mainstream science fiction. Fiction can often serve as inspiration for speculative design and scientific exploration, leading to tangible change in the real world. I hope that this essay has provoked thoughts and interest in these topics, and can serve as a starting point for discussions on aspirational queer robotics futures.

Most importantly, an aspirational future should include community and solidarity among queer roboticists. \cite{stolp2024more} advocate for the recognition of queer people as robot designers and researchers. This is already taking place. Queer in Robotics\footnote{\url{https://sites.google.com/view/queerinrobotics/}, Accessed 6th June 2025} is a community of queer roboticists, holding gatherings at robotics conferences such as the Human-Robot Interaction conference (known as ``HRI'', published by ACM/IEEE) and the International Conference on Robotics and Automation (known as ``ICRA'', published by IEEE). Whatever the future of robot identity, a queer robotics future should include spaces where queer roboticists can support and celebrate each other. 

\section{Presentation Format for the Conference}

This paper will be presented in a poster format, and will include participatory generative ideation for queer robotics futures. To do this, I will present some of the questions posed in this essay on a board. I will engage with conference participants in discussions about each of these questions, and encourage them (with their consent) to write down their thoughts on post-it notes, and include them on the poster. Participants will also be invited to sketch or write down their speculative design visions for aspirational queer robotics futures. The result will be a collaborative sense-making board of queer robotics futures, which will evolve throughout the presentation session. This will enable participants to engage with the thought-provoking questions of the essay, and visually see each others' perspectives and visions of the future.

\section{Acknowledgements}

M. A. acknowledges a postdoctoral fellowship grant from the Emil Aaltonen Foundation. Thank you to Robin and Lux for thought-provoking discussions related to this manuscript.

\end{document}